\documentclass[letterpaper, 10 pt, conference]{ieeeconf}
\IEEEoverridecommandlockouts
\let\labelindent\relax
\let\labelindent\relax
\usepackage{graphicx} \graphicspath{ {figures/} }
\usepackage{amsmath,amssymb,mathabx,amsbsy,mathtools,etoolbox}
\usepackage[utf8]{inputenc} 
\usepackage[T1]{fontenc}    
\usepackage{url}
\usepackage{algorithmic}
\usepackage[ruled]{algorithm2e}

\usepackage{acronym}
\usepackage{enumitem}
\usepackage{balance}
\usepackage{xspace}
\usepackage{setspace}
\usepackage[skip=3pt,font=small]{subcaption}
\usepackage[skip=3pt,font=small]{caption}
\usepackage[dvipsnames,svgnames,x11names]{xcolor}
\usepackage[capitalise,nameinlink]{cleveref}
\usepackage{booktabs,tabularx,colortbl,multirow,multicol,array,makecell}
\usepackage{pifont}
\usepackage{cite}
\usepackage{nicematrix}
\usepackage{dblfloatfix}

\makeatletter
\DeclareRobustCommand\onedot{\futurelet\@let@token\@onedot}
\def\@onedot{\ifx\@let@token.\else.\null\fi\xspace}

\def\etc{\emph{etc}\onedot}

\def\etal{\emph{et al}\onedot}
\makeatother

\makeatletter
\def\BState{\State\hskip-\ALG@thistlm}
\makeatother

\makeatletter
\renewcommand{\paragraph}{%
  \@startsection{paragraph}{4}%
  {\z@}{0ex \@plus 0ex \@minus 0ex}{-1em}%
  {\hskip\parindent\normalfont\normalsize\bfseries}%
}
\makeatother

\crefname{algorithm}{Alg.}{Algs.}
\Crefname{algocf}{Algorithm}{Algorithms}
\crefname{section}{Sec.}{Secs.}
\Crefname{section}{Section}{Sections}
\crefname{table}{Tab.}{Tabs.}
\Crefname{table}{Table}{Tables}
\crefname{figure}{Fig.}{Fig.}
\Crefname{figure}{Figure}{Figure}

\definecolor{gblue}{HTML}{4285F4}
\definecolor{gred}{HTML}{DB4437}
\definecolor{ggreen}{HTML}{0F9D58}

\definecolor{mygray}{gray}{.92}

\definecolor{lightgray}{gray}{0.9}

\acrodef{vbts}[VBTS]{Vision-based Tactile Sensor}
\acrodef{qp}[QP]{Quadratic Programming}
\acrodef{dof}[DoF]{Degree of Freedom}
\acrodef{ros}[ROS]{Robot Operating System}
\acrodef{uav}[UAV]{Unmanned Aerial Vehicle}
\acrodef{dof}[DoF]{Degree of Freedom}
\acrodef{com}[CoM]{center of mass}
\acrodef{msrr}[MSRR]{modular self-reconfigurable robot}
\acrodef{owmr}[OWMR]{Omnidirectional Wheeled Mobile Robot}
\acrodef{wmr}[WMR]{Wheeled Mobile Robot}
\acrodef{ftac}[F-TAC Hand]{Full-hand TACtile-embedded anthropomorphic Hand}
\acrodef{gep}[GEP]{Generalized Earley Parser}
\acrodef{t-aog}[T-AOG]{temporal And-Or Graph}
\acrodef{dps}[DPS]{Deep Photometric Stereo}
\acrodef{mala}[MALA]{Metropolis-Adjusted Langevin Algorithm}
\acrodef{mcp}[MCP]{metacarpophalangeal}
\acrodef{pip}[PIP]{proximal interphalangeal}
\acrodef{dip}[DIP]{distal interphalangeal}
\acrodef{adelm}[ADELM]{Attraction-Diffusion Energy Landscape Mapping}
\acrodef{dof}[DoF]{Degree-of-Freedom}
\acrodef{meps}[MEPs]{Minimum Energy Paths}
\acrodef{pca}[PCA]{Principle Component Analysis}
\acrodef{svc}[SVC]{Support Vector Classifier}
\acrodef{fpsampling}[FPS]{Furthest Point Sampling}
\acrodef{t-SNE}[t-SNE]{t-distributed Stochastic Neighbour Embedding}
\acrodef{rbf}[RBF]{Radial Basis Function}

\title{\LARGE \bf  Image Acquisition, Design and Calibration of Vision-based Tactile Sensor for Large-scale Deployment on Multi-fingered Grippers}
\title{\LARGE \bf Large-scale Deployment of Vision-based Tactile Sensors \\ on Multi-fingered Grippers}

\author{Meng Wang$^{1*}$, Wanlin Li$^{1*}$, Hao Liang$^{1}$, Boren Li$^{1}$, Kaspar Althoefer$^{2}$, Yao Su$^{1\dagger}$, and Hangxin Liu$^{1\dagger}$ 
\thanks{*Meng Wang and Wanlin Li contributed equally to this work.}
\thanks{$\dagger$ Corresponding authors.}
\thanks{$^{1}$ State Key Laboratory of General Artificial Intelligence, Beijing Institute for General Artificial Intelligence (BIGAI). Emails: \tt{\{wangmeng, liwanlin, lianghao, liboren, suyao, liuhx\}@bigai.ai}}
\thanks{$^{2}$ Centre for Advanced Robotics @ Queen Mary (ARQ), Queen Mary University of London. Email: \tt{k.althoefer@qmul.ac.uk}.}
}

\begin{document}

\maketitle   

\begin{abstract}
\acp{vbts} show significant promise in that they can leverage image measurements to provide high-spatial-resolution human-like performance. However, current \ac{vbts} designs, typically confined to the fingertips of robotic grippers, prove somewhat inadequate, as many grasping and manipulation tasks require multiple contact points with the object. With an end goal of enabling large-scale, multi-surface tactile sensing via \acp{vbts}, our research (i) develops a synchronized image acquisition system with minimal latency, (ii) proposes a modularized \ac{vbts} design for easy integration into finger phalanges, and (iii) devises a zero-shot calibration approach to improve data efficiency in the simultaneous calibration of multiple \acp{vbts}. In validating the system within a miniature 3-fingered robotic gripper equipped with 7 \acp{vbts} we demonstrate improved tactile perception performance by covering the contact surfaces of both gripper fingers and palm. Additionally, we show that our \ac{vbts} design can be seamlessly integrated into various end-effector morphologies significantly reducing the data requirements for calibration.
\end{abstract}

\section{Introduction}
Humans exhibit remarkable adaptability in grasping and manipulation tasks, ranging from power-gripping heavy tools to delicate handling of fragile items under a variety of conditions. Robotic devices have, thus far, fallen short in terms of achieving these human-like levels of adaptability. Given the crucial role of tactile perception in human hands' functionality, as various studies have indicated~\cite{johansson20086,jenmalm2003influence,schneider2016anticipation,flanagan2006control}, conferring a sense of touch could be the key ingredient. Indeed, the utilization of tactile feedback has significantly advanced robots' manipulation skills in bottle cap opening~\cite{edmonds2019tale}, handling fragile objects~\cite{li20233}, object reorientation~\cite{yin2023rotating}, cable manipulation~\cite{she2021cable} \etc. However, unlike humans who use the entirety of the hand to sense and grasp, object manipulation in robots is based on tactile sensing that is principally restricted to its fingertips. Large-scale, multi-surface integration of tactile feedback into multi-fingered grippers remains a pivotal challenge due to the complex articulation of these mechanical structures. Finding an effective solution to this issue could, however, radically improve flexibility and robustness in robotic grasping.

\begin{figure}[t!]
    \centering
    \includegraphics[width=\linewidth,trim=0cm 0cm 0cm 0cm, clip]{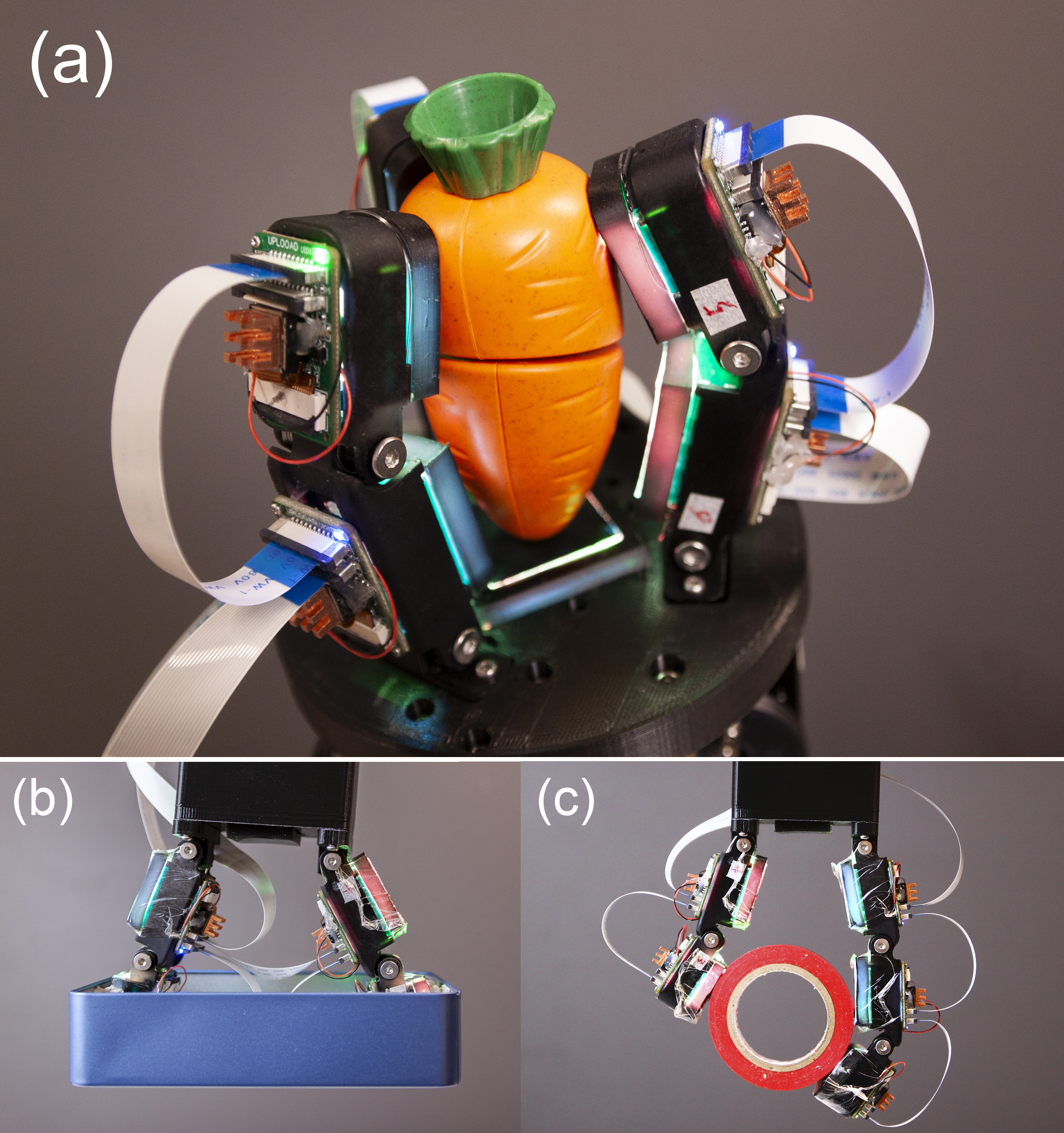}
    \caption{\textbf{Large-scale \acp{vbts} deployment on a robotic gripper.} (a) Acquiring rich tactile information utilizing the three-fingered robotic gripper. Seven modular \acp{vbts} are deployed into the phalanges of each finger as well as the palm of the gripper. (b)(c) Alternative configurations using modular \acp{vbts} for other tasks. }
    \label{fig:motiv}
\end{figure}

The prevalent method of incorporating large-scale, multi-surface tactile sensing into standard robotic grippers is the \textit{augmentation} approach, whereby a dedicated layer of devices, such as force-sensitive resistors~\cite{yin2023rotating}, piezoresistive materials~\cite{liu2017glove}, strain gauges~\cite{ATISensor}, or e-skin~\cite{park2015fingertip,yu2022all}, is affixed to the surface of the gripper. However, these setups may well be too rigid to conform to the contours of a multi-fingered gripper, hindering fine manipulative tasks. Additionally, discrepancies between the mounted sensors and the underlying mechanical structure could lead to inaccuracies, reminiscent of the skin-motion artifact observed in human movement. A more effective alternative is the \textit{embedding} approach, in which sensors are integrated into the mechanical structure of the gripper. Typically, these might include high-fidelity sensors like BioTact~\cite{fishel2012sensing}, force transducers, and low-cost \acp{vbts}, such as GelSight~\cite{yuan2017gelsight} or its variants~\cite{donlon2018gelslim,li2020f,wang2021gelsight,li20233}. 

Of all robotic tactile sensor types, \acp{vbts} are deemed the ideal candidates for large-scale deployment on multi-fingered robotic grippers. Using just images, they can acquire pixel-level tactile resolution comparable to human skin - a method significantly easier than constructing sensing arrays like those in e-skin or MEMS transducers -  and have contact modules that are elastic and can be easily moulded into different shapes~\cite{donlon2018gelslim,li2020f,wang2021gelsight,li20233} to suit various gripper structures and target object properties. However, a key challenge remains, relating to their bulkiness, despite the potential availability of slimmer versions~\cite{ma2019dense,lambeta2020digit,wang2021gelsight}.
In addition to this, \textbf{data synchronization}, \textbf{easy integration}, and \textbf{efficient calibration} are three further difficulties associated with a system involving multiple \acp{vbts}.

In this research, we aim to address these issues and endow a multi-fingered robotic gripper with human-like tactile perception performance by integrating multiple modular \acp{vbts} into each phalange as well as into the palm. To achieve this goal, we first propose a synchronized image acquisition system to mitigate the drawbacks of USB/CSI cameras which are typically used in \acp{vbts}. Then we introduce a modular tactile sensor design that minimizes dimensions, weight, and cabling to ease its integration into finger phalanges and other gripper morphologies. We also develop a zero-shot approach to efficient calibration across multiple sensors. Finally, we validate the entire system creating a 3-fingered robotic gripper in which seven modular tactile sensors operate simultaneously and in which tactile perception is widely achievable across the contact surfaces of both gripper fingers and palm.  

\subsection{Related Work}

\textbf{Large-scale Tactile Perception:} The field of large-scale tactile perception has seen significant advancements through innovative sensor technologies~\cite{hammock201325th}. Various techniques, using piezoresistive sensors~\cite{liu2017glove}, capacitive sensors~\cite{ji2021gradient}, piezoelectric sensors~\cite{lin2021skin}, magnetic sensors~\cite{yan2021soft}, optical sensors~\cite{zhang2022deltact}, \etc, have been explored to provide robotic end-effectors with extensive tactile feedback for object assessment via physical contact~\cite{ahanat2015tactile}. Key design parameters such as spatial resolution, sensitivity, wiring complexity, frequency response, flexibility, robustness, and cost are crucial. Balancing these criteria often requires trade-offs in practical applications. While e-skins~\cite{park2015fingertip,yu2022all} perform relatively well in the aforementioned categories, they still face challenges such as manufacturing complexity and high cost, as well as having wiring and durability issues.

\textbf{GelSight Variants:}  \acp{vbts} demonstrate advantages over alternatives in terms of high spatial resolution~\cite{yuan2017gelsight,ward2018tactip}, compliance~\cite{alspach2019soft}, durability~\cite{lambeta2020digit}, low-cost~\cite{abad2020low,lin20239dtact} and wireless~\cite{li20233} characteristics, and, in recent years, several attempts have been made to deploy multiple such \acp{vbts} in robotic end-effectors. For instance, Padmanabha \etal~\cite{padmanabha2020omnitact} utilize three micro-cameras to realize a multi-directional high-resolution tactile sensor; Sferrazza \etal~\cite{trueeb2020towards} describe a multi-camera tactile sensor that shows promise for vision-based robotic skins; Liu \etal~\cite{liu2023gelsight} describe a soft endoskeleton three-fingered robot hand that extracts rich tactile data during grasping; Wilson \etal~\cite{wilson2020design} present a two-fingered robot gripper with multiple GelSight sensors that can determine the surface topology of the object; She \etal~\cite{she2020exoskeleton} describe a vision-based proprioceptive and tactile sensing finger for soft robots; Zhao \etal~\cite{zhao2023gelsight} present a vision-based three-fingered robotic hand that obtains rich tactile signal during grasping; Romero \etal~\cite{romero2020soft} demonstrate rounded GelSight fingertips that can be installed on an Allegro hand to capture contact surfaces in high-resolution. However, as summarized in \cref{tab:bom}, these attempts present the simultaneous operation of no more than four sensors and lack a synchronizing mechanism to support a large increase in the number of sensors. With this in mind, we try to enhance the design of \ac{vbts} to facilitate their large-scale deployment in robotic grippers. A novel three-fingered robotic gripper that incorporates seven \acp{vbts} is presented, in which the entire inner surface of both fingers and palm are imbued with tactile perception, and in which all  \acp{vbts} operate simultaneously. 

\begin{table}[t!]
    \caption{\textbf{The current large-scale vision-based tactile systems.}}
    \centering
    \label{tab:bom}
    \resizebox{0.95\linewidth}{!}{%
        \rowcolors{1}{white}{lightgray}
        \begin{tabular}{lccc}
        \toprule
        & \textbf{Name} & {\makecell[c]{\textbf{Maximum No. of sensors} \\ \textbf{working simultaneously}}} & {\makecell[c]{\textbf{No. of sensors} \\ \textbf{equipped}}} \\
        \hline
        & Omnitact~\cite{padmanabha2020omnitact} & $3$   &  $3$\\
        & Multi-camera Tactile Sensor~\cite{trueeb2020towards} & $4$ & $4$\\
        & GelSight EndoFlex~\cite{liu2023gelsight} & $4$ & $6$\\
        & {\makecell[c]{Fully Actuated Robotic Hand \\ With Multiple Gelsight~\cite{liu2023gelsight}}} & $4$ & $4$\\
        & {\makecell[c]{Vision-based Exoskeleton-covered \\ Soft Finger~\cite{she2020exoskeleton}}} & $4$ & $4$\\
        & GelSight Svelte Hand~\cite{zhao2023gelsight} & $3$ & $3$\\
        & {\makecell[c]{Soft, Round, and High-Resolution \\ Soft Finger~\cite{romero2020soft}}} & $4$ & $4$\\
        \hline
        \end{tabular}
        }%
\end{table}

\subsection{Overview}
We organize the remainder of the paper as follows. \cref{sec:design} outlines the hardware design of the proposed modular tactile sensor. \cref{sec:cali_formation} describes the zero-shot calibration method while \cref{sec:deploy} presents its deployment on a three-fingered robotic gripper. We discuss and conclude the paper in \cref{sec:discussion,sec:conclusion}.

\section{Modular Tactile Sensor Design}\label{sec:design}
\subsection{Synchronized Image Acquisition System}\label{sec:system_design}

\begin{figure}[b!]
    \centering
    \includegraphics[width=\linewidth,trim=0cm 0.8cm 0cm 0cm, clip]{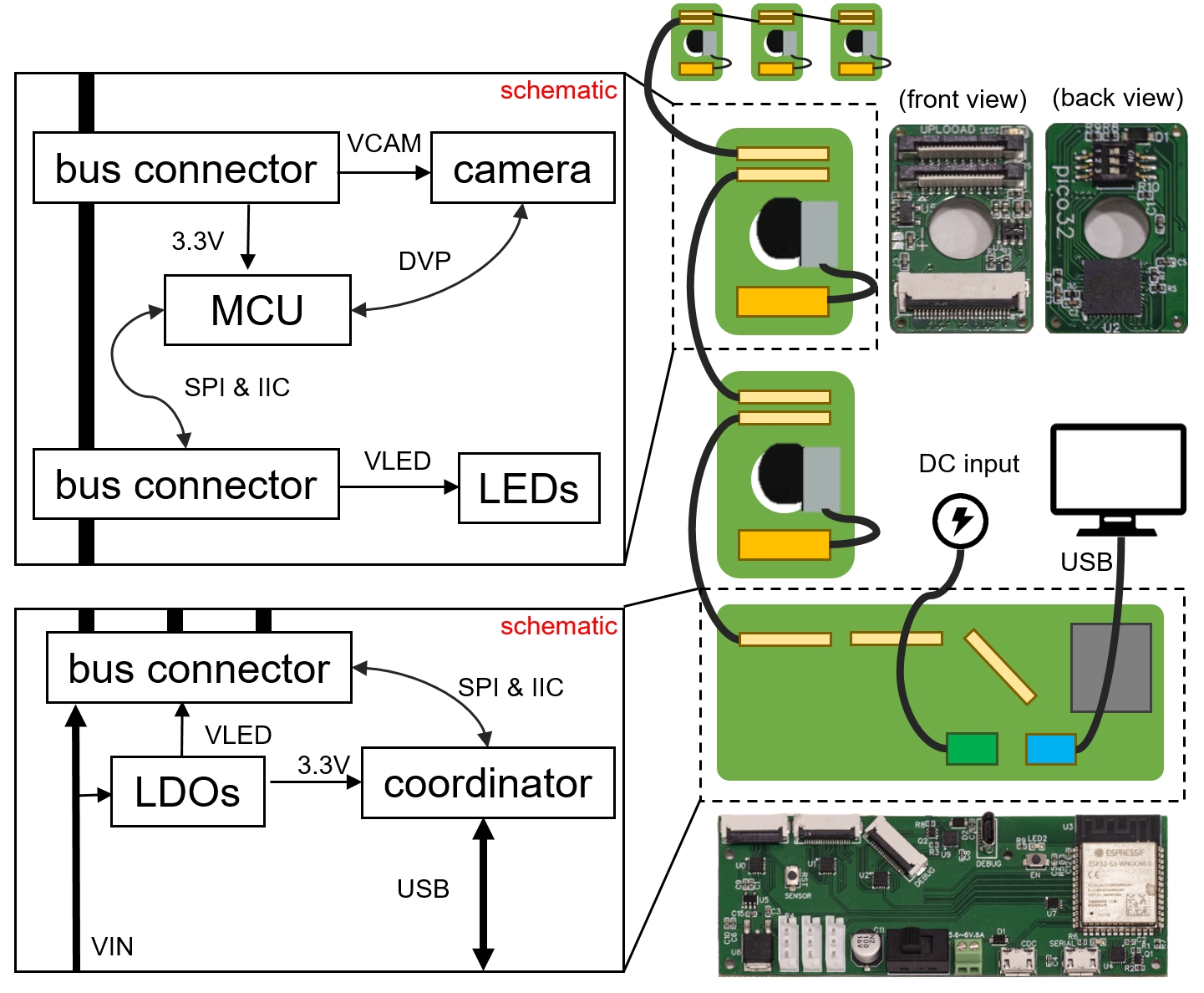}
    \caption{\textbf{Synchronized image acquisition system.} Multiple sensor boards are cascade-connected to the hub board through FFC cables. The hub board provides the power supply for both the sensor boards and the motors, and handles the communication between the acquisition system and the host PC.}
    \label{fig:system}
\end{figure}

Traditional GelSight sensors, which utilize USB or CSI cameras to capture vision-based tactile information, pose challenges for deployment across multiple sensing planes due to their large dimensions and separate wiring. Moreover, these sensors often lack temporal synchronization in image acquisition, a problem that intensifies as the number of sensors increases. To address these drawbacks, we propose a modular image acquisition system specifically designed to incorporate multiple image sensors within the confined spaces of finger-like actuators.

The acquisition system (see \cref{fig:system}) comprises a hub board and multiple sensor boards that can be cascade-connected via FFC cables or customized FPCs. Each sensor board is equipped with an MCU and communicates with a miniature camera through a DVP port. The hub board provides assorted connectors for linking sensor boards, motors, power supply, and the host computer. It also features a core coordinator, which can be either a USB multifunction converter IC (e.g., CH347) or an RTOS-supported MCU for greater flexibility. The coordinator manages the communication bus, which includes an SPI bus for image acquisition and an IIC bus for sensor control. Power modules like LDOs are also placed on the hub board to reduce heat of sensors. 

\textbf{Temporal Synchronization Design:} To enhance synchronization accuracy, we implement an additional control signal that dictates the precise moment for image capture by the sensors. Operating on a Fast-mode I2C bus at a frequency of $400~\mathrm{kHz}$, the time required for each sensor to capture an image is remarkably less than $30~\mu s$, ensuring high synchronization. 

\begin{figure}[t!]
\centering
\includegraphics[width=\linewidth,trim=0cm 0cm 0cm 0cm, clip]{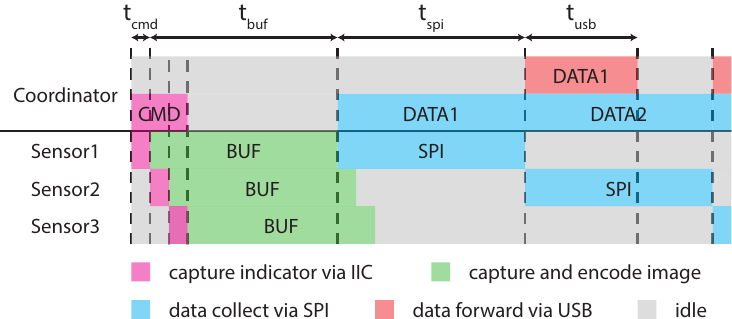}
\caption{\textbf{Sequence diagram of the acquisition process.} The system is designed to minimize synchronization error, and thus provide a high frame rate with limited latency.}
\label{fig:synchronization}
\end{figure}

\textbf{Theoretical Performance:} Image acquisition commences immediately upon readiness of the first sensor. The coordinator sequentially collects images from the sensors via the SPI bus and forwards them to the host computer via USB. The system performance heavily depends on the number of sensors $N$, and the required image frame size $S$. Given that the USB bitrate typically surpasses the SPI bitrate, the system's overall frame rate is primarily determined by the latter. 
To optimize frame rate and minimize latency, the SPI bus operates at a $40~\mathrm{MHz}$ clock frequency in DMA mode. With a common frame size (320x240, JPEG) of $S=20~\mathrm{KB}$, the transmission time for every image is $t_{\text{spi}}=4~\mathrm{ms}$, while the initial image buffering time $t_{\text{buf}}$ is approximately $1~\mathrm{ms}$. As such, the frame rate is computed as $f=1/(Nt_{\text{spi}}+t_{\text{buf}})$, with the maximum latency being $\text{latency}=Nt_{\text{spi}}+t_{\text{buf}}+t_{\text{usb}}$. The maximum synchronization error is calculated as $\text{error}_{\text{sync}}=30~$N$\mu$s. This synchronization timeline is illustrated in \cref{fig:synchronization}.

\setstretch{0.98}
\subsection{Modular Tactile Sensor Design \& Fabrication}\label{sec:sensor_design}

\begin{figure}[t!]
\centering
\includegraphics[width=\linewidth,trim=0cm 1cm 0cm 0cm, clip]{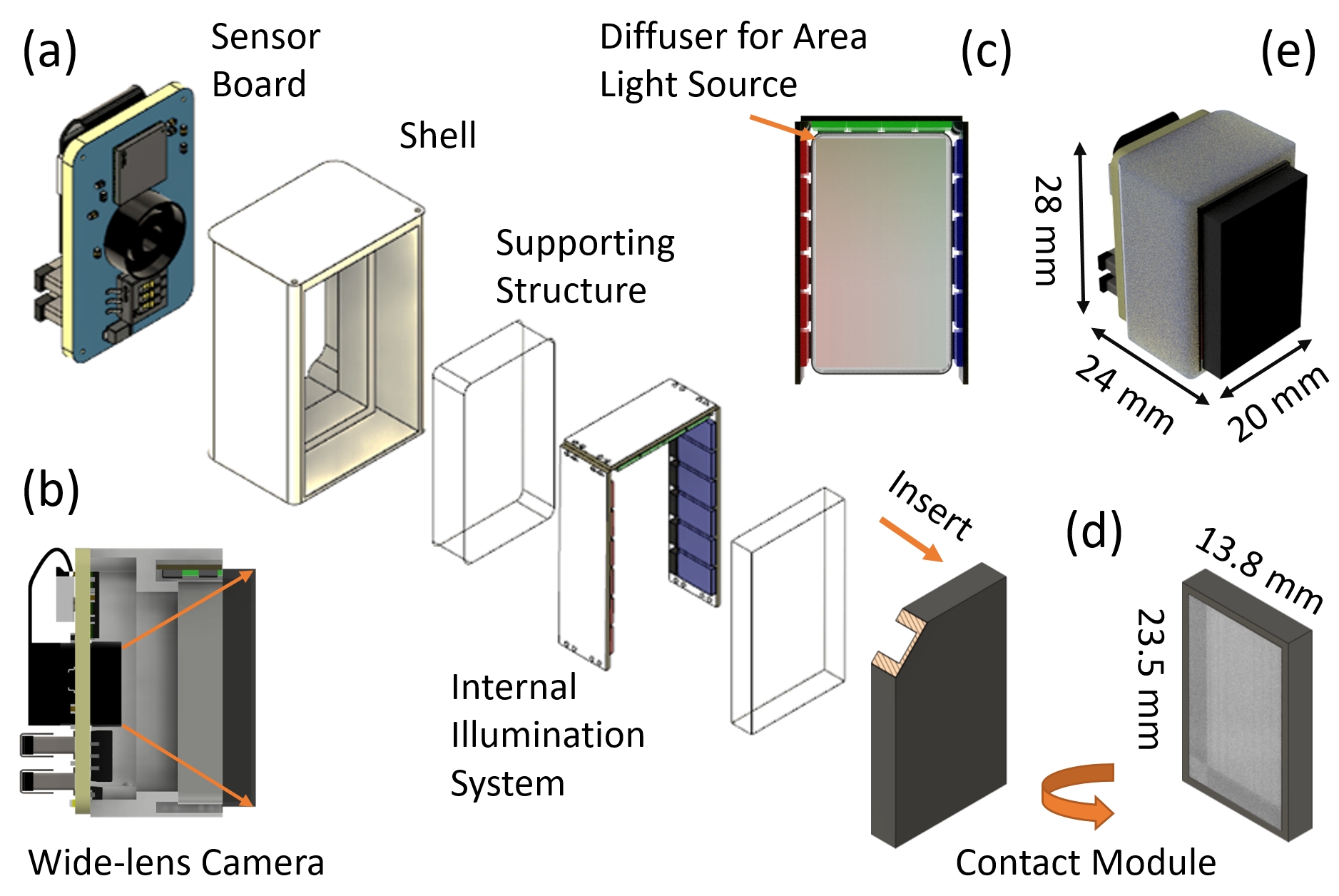}
\caption{\textbf{Design and fabrication of the modular vision-based tactile sensor.}  (a) The structure of the \ac{vbts} consists of a modular image acquisition board, a wide-lens camera, an illumination system, a supporting structure, a contact module, and a shell. (b) The view of the wide-lens camera covers the whole surface of the contact module. (c) The internal area lighting system utilizes side illumination towards the supporting structure for shade accentuation and compactness. (d) The elastomer serves as the contact module to provide super-high spatial resolution. (e) The overall dimensions of each sensor are minimized to $28 \times 20 \times 19~\mathrm{mm}$.}
\label{fig:fab}
\end{figure} 

The dimensions of \acp{vbts} are primarily determined by the image acquisition system. Leveraging the modular system presented above, we can achieve a compact design. As illustrated in \cref{fig:fab}(a), the dimensions of the sensor are minimized to $28 \times 20 \times 19~\mathrm{mm}$ and its weight to $20~g$ - figures comparable to those of a human phalanx. The proposed sensor comprises a modular image acquisition board, a wide-lens camera, an internal illumination system, a supporting structure, a contact module, and a shell, as depicted in \cref{fig:fab}(b). The fabrication process of each component is detailed below. 

\begin{figure*}[t!]
\centering
\includegraphics[width=\linewidth,trim=0cm 0.5cm 1cm 0cm, clip]{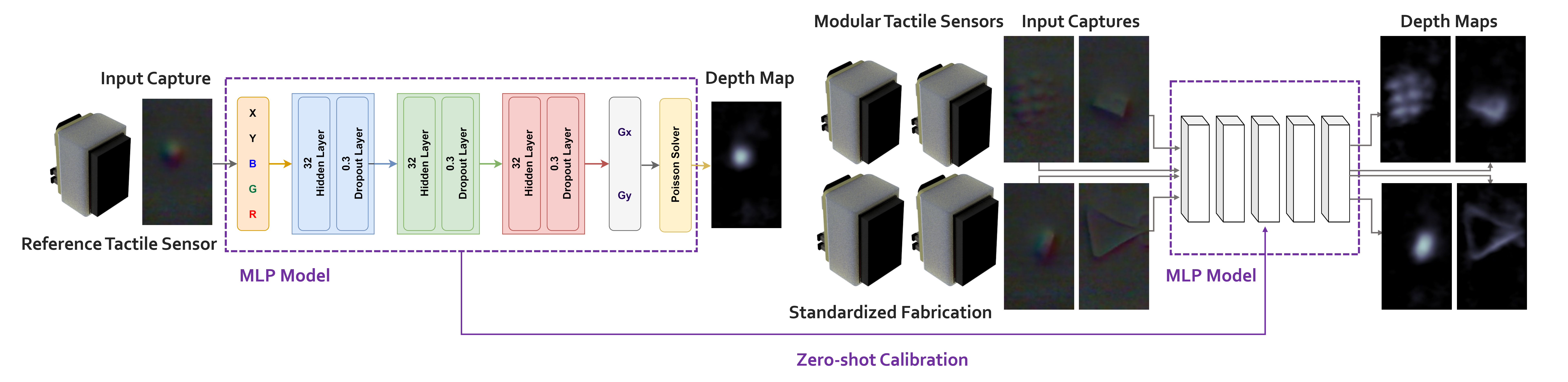}
\caption{\textbf{The zero-short calibration process for large-scale deployment of \acp{vbts}.}  To jointly calibrate the \acp{vbts} that share the similiar configuration, a multiple-layer perception (MLP) neural network is utilized to learn the mapping between pixel intensity within RGB color channels, position information, and gradients in the x and y directions.}
\label{fig:calibration}
\end{figure*} 

\textbf{Wide-lens Camera:} As illustrated in \cref{fig:fab}(c), we used an OmniVision OV2640 CMOS camera module with a 160-degree viewing lens to provide an open view of $2$ megapixels. Apart from camera calibration to counter wide-lens distortion, parameters such as exposure and white balance are manually configured to align with the internal illumination system and ensure optimal sensor performance.

\textbf{Modular Image Acquisition Board:} The sensor board accommodates the camera lens in a central aperture, resulting in a compact assembly. An onboard MCU captures and stores images, while two external connectors facilitate bus cascading. In addition, separate power is supplied to the illumination system to provide stable lighting conditions.

\textbf{Internal Illumination System:} Amongst several illumination configurations in the design of \acp{vbts}, illumination towards the supporting structure is generally used in flat, compact GelSight sensors~\cite{wang2021gelsight}, operating under the principle of grazing lighting that accentuates surface texture by exaggerating its shades. To achieve this effect, we use multiple flat-surface 3528 SMD LEDs with a light diffuser, creating an area surface lighting source that directly illuminates the side of the supporting structure, as depicted in \cref{fig:fab}(d). Therefore, any deformation of the contact module leads to a significant change in the optical path, visible to the camera.

\textbf{Supporting Structure:} The supporting structure is in direct contact with the underside of the soft contact module and facilitates the camera's observation of the deformations on the other side of the module. Rigidity and transparency are therefore key requirements. For ease of self-crafting, optional materials include acrylics (sheet, tube, shells), epoxy resin (reactive prepolymers), and clear resin (Stereolithography 3D printing). Here, we chose a $5~\mathrm{mm}$ thickness acrylic sheet (in accordance with the LED's dimension) and customized its shape with a laser cutter.

\textbf{Contact Module:} \acp{vbts} typically have elastomers as their contact modules, which can be handcrafted~\cite{yuan2017gelsight} or 3D-printed~\cite{ward2018tactip}. Handcrafted elastomers offer modular fabrication with finer surfaces, enabling higher spatial resolution. In contrast, 3D printing typically generates dot patterns. In this study, we utilized the handcrafted approach as per the Gelsight sensor in~\cite{yuan2017gelsight}, in which the contact module comprises a transparent silicone base and a coating layer with a Lambertian surface. The transparent silicone base is made of Part A$\&$B and silicone thinner Slacker in a weighted ratio of $1:1:3$ which produces a moderate sensing medium. The thickness of the base is fabricated at $3~\mathrm{mm}$. We airbrushed the diluted mixture of spherical aluminum powder, silicone solvent, and Smooth-On NOVOCS Matte (weighted ratio of $1:10:30$) onto the base's upperside to form a reflective coating layer with a Lambertian surface. Finally, we added a thin protective layer made of mill-resistant matte oil to increase the sensor's durability. The final contact module is depicted in \cref{fig:fab}(e)

\textbf{Shell:} The shell is essential for securing each component within the sensor while preventing external environmental light from affecting the internal capture system. To achieve this, we utilized Stereolithography (SLA) 3D printing technology to produce the shell from lightweight black resin material. The manufactured shell displays good rigidity and toughness, while effectively blocking ambient light.

\section{Modular Sensor Calibration}\label{sec:cali_formation}
In this section, we describe a zero-shot calibration approach to the large-scale deployment of the proposed \acp{vbts}, which have identical dimensions and setups on each phalange of a robotic multi-fingered gripper. We then present a performance comparison between the proposed calibration approach and individual calibration for multiple sensors.

The principle of GelSight-like tactile sensors is based on photometric stereo which maps the captured pixel intensity to the change in surface gradient~\cite{yuan2017gelsight}. A fast Poisson solver is then applied to integrate gradients to obtain the depth map. Due to its modular design, each sensor module shares almost the same configuration, including identical dimensions, illuminating systems, camera settings, and contact module composite. Though few variations might introduced during
fabrication and assembly, it still makes zero-shot calibration feasible with fine-tuning techniques.

In this study, we employ a multiple-layer perception (MLP) neural network to learn the mapping between pixel intensity within three color channels, position information, and gradients in the $x$ and $y$ directions, as depicted in \cref{fig:calibration}. The network is composed of three hidden layers and three dropout layers with the $Tanh$ activation function. The training dataset for a single sensor consists of 50 captures, and the testing dataset comprises $5$ captures. We conducted trials using both the raw capture and the differential capture (between current frame and reference frame) as inputs to the model. When transitioning to another modular \ac{vbts}, we initially fine-tune a single difference capture to achieve alignment with the reference sensor, by applying a constant offset to each color channel. Then we directly apply the model to achieve automatic calibration.

\begin{table}[b!]
\small
    \caption{\textbf{Evaluation of proposed calibration approaches on three modular tactile sensors using look-up table and MLP}}
    \centering
    \label{tab:calibration}
    \resizebox{1\linewidth}{!}{%
        \begin{tabular}{c|ccccc}
        \hline
        \multirow{3}{*}{\textbf{Groups}} & \multicolumn{5}{c}{\textbf{Calibration Approaches}} \\ 
        \cline{2-6}
                  & \makecell[c]{Individual \\ Lookup Table}
                  & \makecell[c]{Individual \\ Raw Input}
                  & \makecell[c]{Individual \\ Diff Input} & \makecell[c]{Zero-shot \\ Raw Input}  & \makecell[c]{Zero-shot \\ Diff Input}  \\
        \hline
        Data Collected & $150$ & $150$  & $150$  &  $50$ &  $50$\\
        \hline
        $G_x$ error (MAE) & $0.041$ & $0.034$  & $0.028$  &  $0.042$ &  $0.034$\\
        \hline
        $G_y$ error (MAE) & $0.048$ & $0.043$  & $0.034$  & $0.051$ &  $0.039$\\
        \hline
        \end{tabular}
        }%
\end{table}

\cref{tab:calibration} presents quantitative results for the reconstruction using different approaches: individual calibration and zero-shot calibration. As can be seen, utilizing the difference input for zero-shot calibration results in errors closer to those of the individual calibration approach as compared to the raw input, while significantly reducing the required dataset size, as well as the calibration time. This proves the effectiveness of the proposed zero-shot calibration approach for rapid large-scale deployment of \acp{vbts}. 

\begin{figure}[t!]
    \centering
    \includegraphics[width=\linewidth,trim=0cm 0cm 1.2cm 0cm, clip]{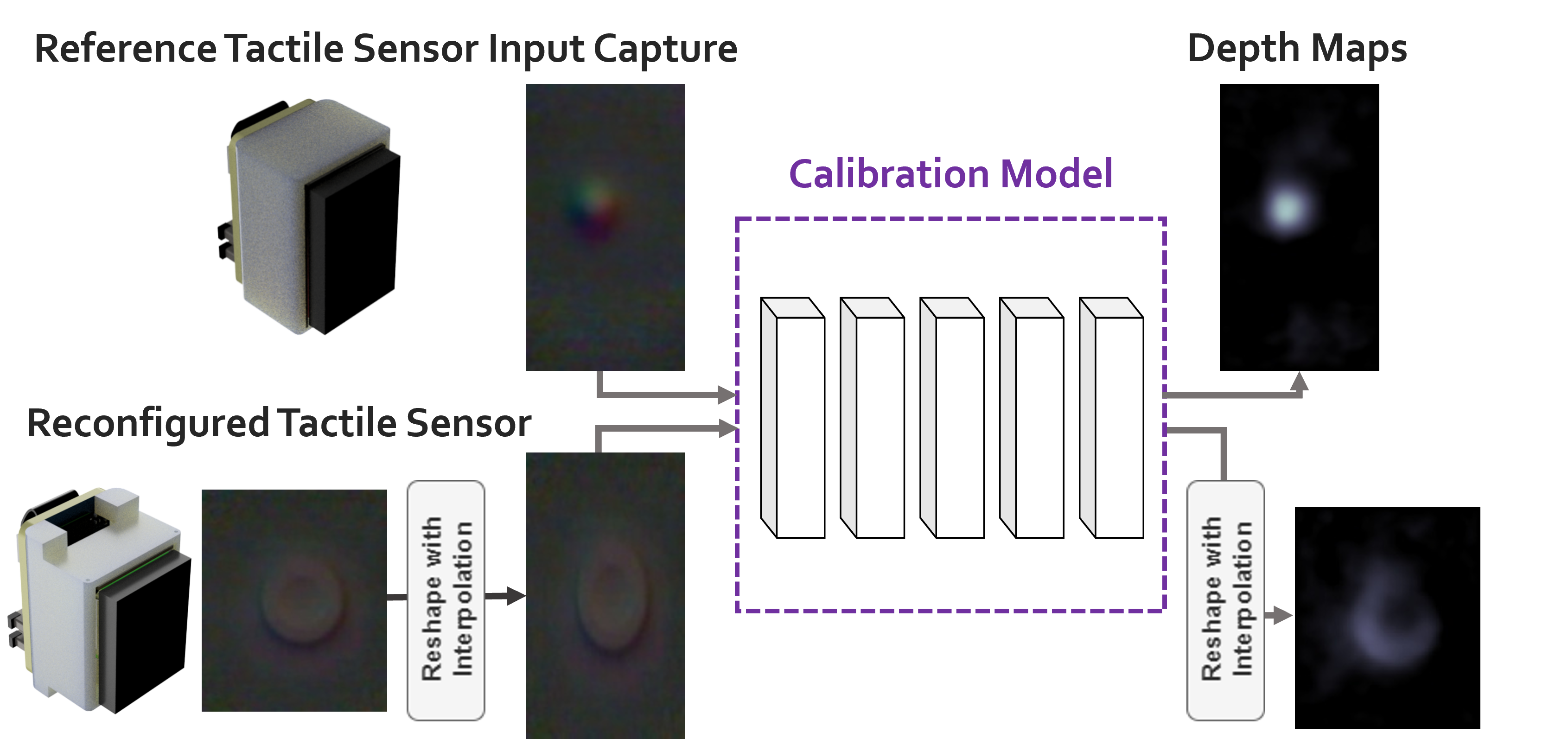}
    \caption{\textbf{The cross-configuration test of the proposed zero-shot calibration model.} Applying the same calibration model on other \acp{vbts} which have non-identical configurations, the model still offers high-fidelity depth map reconstruction performance.}
    \label{fig:cross-config}
\end{figure}

Additionally, we conducted an experiment on \acp{vbts} with similar but non-identical configurations. The qualitative results, as depicted in \cref{fig:cross-config}, demonstrate that despite some differences in size between the two sensors, the model can still reconstruct the depth map in high-fidelity taking advantage of the proposed zero-shot calibration method,  further underscoring its robustness.

\section{Real Deployment on GelGripper} \label{sec:deploy}

\subsection{GelGripper Design} 
\begin{figure}[t!]
    \centering
    \includegraphics[width=0.9\linewidth,trim=0cm 0cm 0cm 1.8cm, clip]{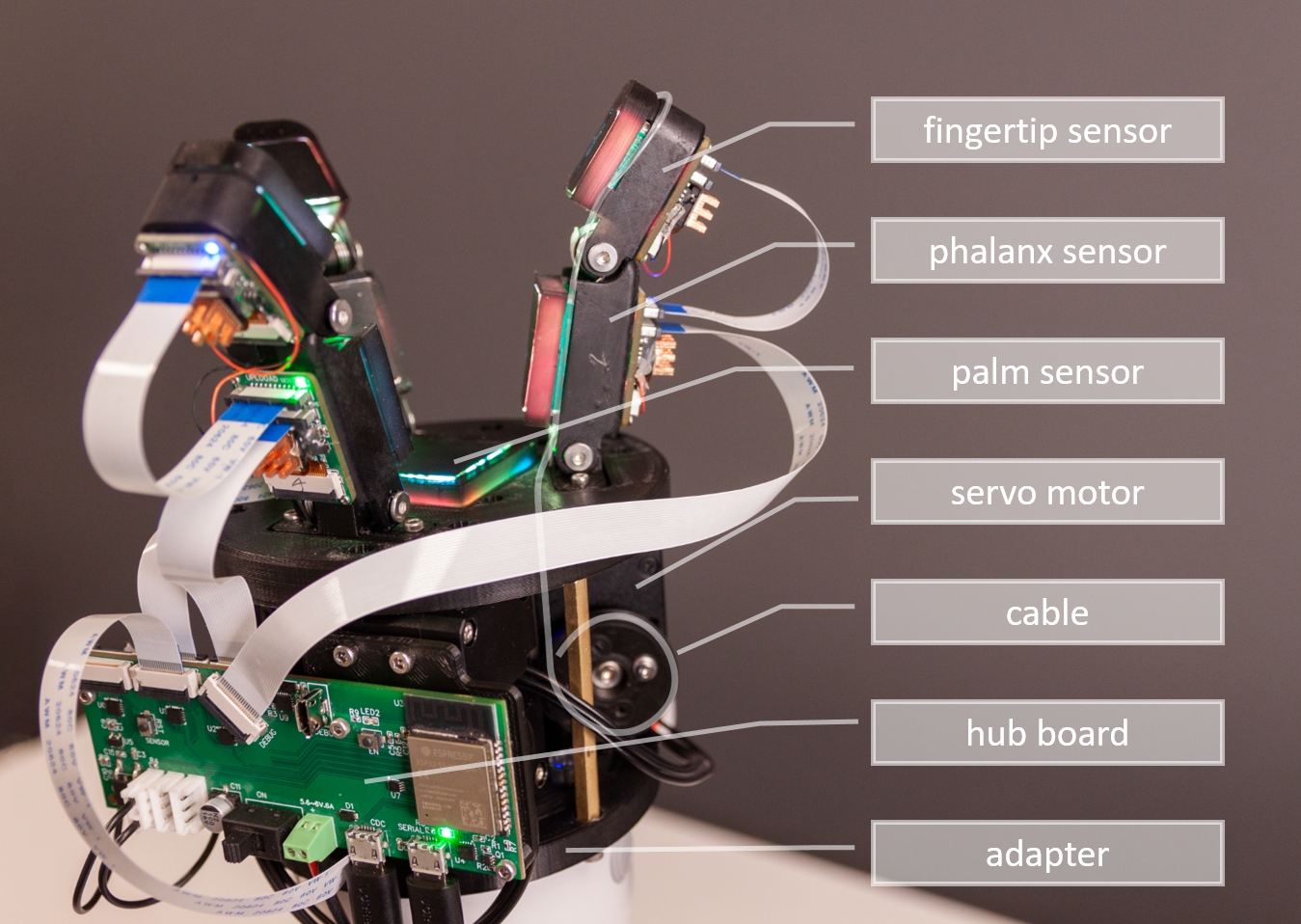}
    \caption{\textbf{The mechanical design of the GelGripper.} Three different types of sensing structures for the fingertip, middle phalanx, and palm are all incorporated within the same sensing hardware. Each finger is actuated by a servo motor through the cable-driven mechanism. The compact design of GelGripper enables seamless integration with robot arms.}
    \label{fig:gripper}
\end{figure}

The modular design enables large-scale deployment of the \ac{vbts}. Here, we present the deployment details on an end-effector named GelGripper, integrating seven of the proposed \acp{vbts}. As illustrated in \cref{fig:gripper}, the GelGripper is a cable-driven three-fingered robotic gripper imbued with tactile sensing capabilities across its inner surfaces, including its palm. Each rigid finger is composed of a pair of embedded modular tactile sensors with springs to form the joints. The three fingers are symmetrically positioned around the palm base, effectively with  $110$ degree gaps between them, while an additional tactile sensor is located in the palm's center. The gripper is able to support seven modular sensors working simultaneously to provide large-scale tactile feedback. Cable-driven actuation is applied to maximize the sensing region available and minimize the weight of the finger section. Three Dynamixel XM430-W350-T servo motors are fitted beneath the palm base to drive the cables and achieve flexion through the finger shells, while finger extension is driven by the spring at each joint. 

\subsection{Grasping with GelGripper} 
\begin{figure}[t!]
    \centering
    \includegraphics[width=\linewidth,trim=0cm 0.7cm 0cm 0.3cm, clip]{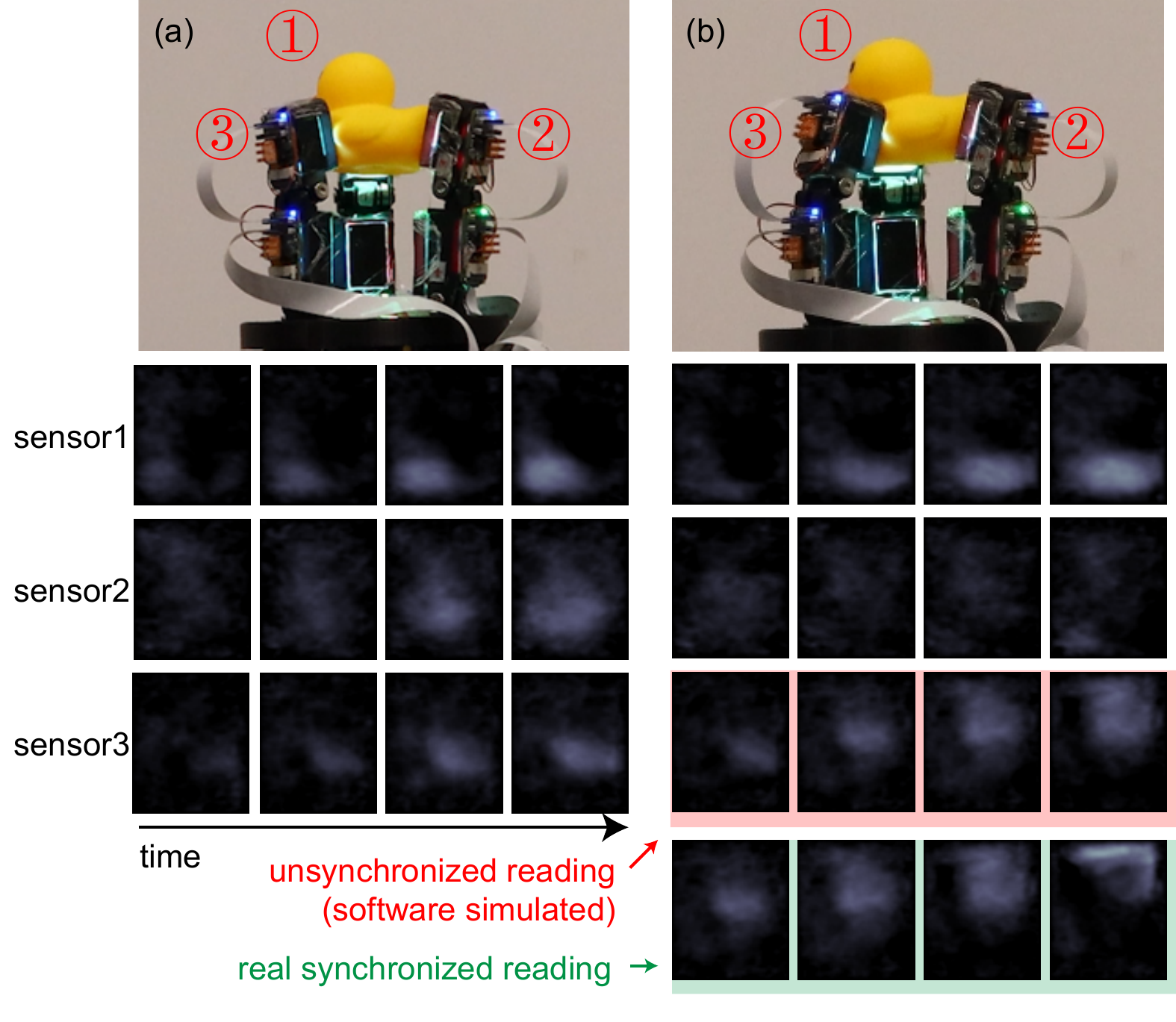}
   \caption{\textbf{Synchronization performance study of \acp{vbts} in gripping.} (a) Gentle handling of a rubber duck achieved with synchronized tactile data. (b) The undesired force that exceeds the threshold appears with unsynchronized tactile data.}
  \label{fig:syn_compare}
\end{figure}

To evaluate the performance of the proposed GelGripper in grasp and manipulation tasks,  we used it to replace the end-effector of a Franka Emika Panda 7-\acp{dof} robot arm with a 3D-printed rigid adapter. As shown in \cref{fig:syn_compare}(a), the sensors can capture subtle changes in the contact imprint in very high resolution, enabling gentle handling of deformable objects. Furthermore, \cref{fig:syn_compare}(b) illustrates a control test in which a software-simulated synchronization error is introduced to one of the sensors. Although the system attempts to achieve a similarly gentle grip, the actual gripping force exceeds the intended level at the point at which the gripper stops, demonstrating the necessity of tactile data synchronization between multiple sensors. 

\begin{figure*}[ht!]
    \centering
    \begin{subfigure}[b]{0.92\linewidth}
        \centering
        \includegraphics[width=\linewidth,trim=0cm 16.3cm 1cm 0cm, clip]{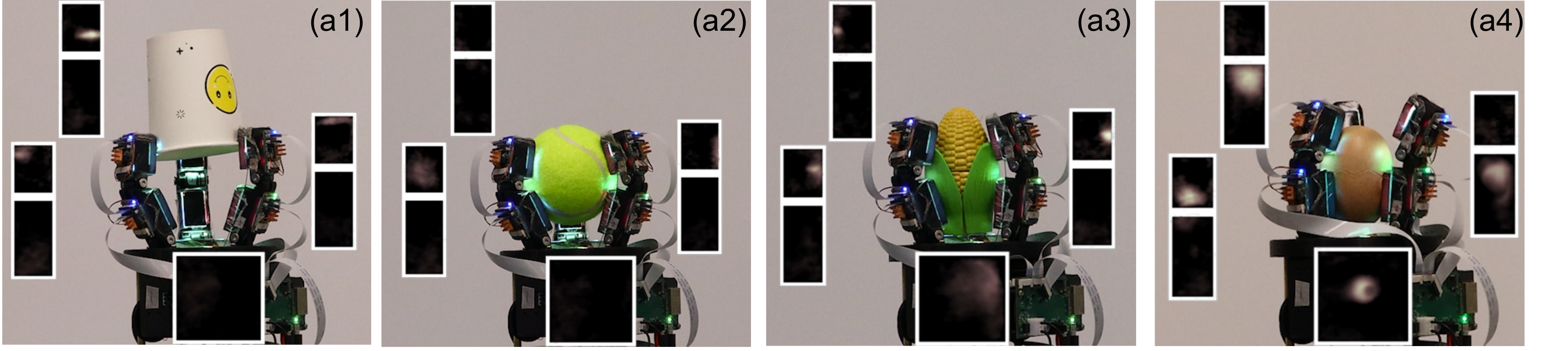}
        \caption{\textbf{Dexterous grasp with four different objects.}}
        \label{fig:grasp}
    \end{subfigure}\\
    \vspace{-5pt}
    \begin{subfigure}[b]{0.92\linewidth}
        \centering
        \includegraphics[width=\linewidth,trim=0cm 0cm 0cm 15.1cm, clip]{figures/data.pdf}
        \vspace{-19pt}
        \caption{\textbf{Forceful manipulation of the screwdriver.}}  \label{fig:screwdriver}
    \end{subfigure}
    \vspace{-10pt}
    \caption{\textbf{Grasp and manipulation with the GelGripper.} (a) Four objects: paper cup (a1), tennis ball (a2), toy corn (a3), egg (a4) are grasped by the GelGripper. (b) The screwdriver is forcefully manipulated by the GelGripper to tighten the bolt with the deformation signal of each sensor plotted in (b4). The whole process is divided into five stages: \textcircled{1} rest status; \textcircled{2} gripping the electric screwdriver body; \textcircled{3} pressing (b1) and starting the trigger (b2); \textcircled{4} tightening the bolt (b3); \textcircled{5} releasing the trigger.
    The imprints of seven \acp{vbts} are also presented in the white boxes within each sub-figure.}
    \label{fig:deploy}
\end{figure*}

 The GelGripper demonstrates its ability to securely grasp everyday objects of varying shapes and levels of hardness. \cref{fig:deploy}(a) illustrates how the GelGripper acquires real-time tactile feedback information from its fingers and palm during these grasping interactions. The GelGripper can also manipulate tools to conduct daily tasks, as demonstrated in \cref{fig:deploy}(b), which illustrates the process of operating an electric screwdriver to tighten a bolt. The entire procedure can be split into five steps: \textcircled{1} at rest; \textcircled{2} secure the body of the screwdriver with the fingers and palm; \textcircled{3} press the trigger of the screwdriver with one of the fingertips until it starts; \textcircled{4} wait for the bolt to become tightened, and \textcircled{5} then release the trigger. \cref{fig:deploy}(b4) illustrates the deformation signal of each sensor throughout the entire procedure.

\section{Discussion}
\label{sec:discussion}

\textbf{Synchronized Image Acquisition:} In contrast to simple parallel grippers, multi-fingered grippers require several sensors operating concurrently to achieve adaptability in grasping. As the number of sensors increases and the frame rate decreases, synchronization becomes essential, to ensure reliable results from subsequent image processing, regardless of fluctuations in frame rate and latency. We developed a prototype in which all sensors were wirelessly connected to facilitate integration with the gripper. However, the prototype struggled to achieve satisfactory synchronization, rendering it sensitive to transmission fluctuations and resulting in poor performance in real-world tasks.

\textbf{Data Transmission and Processing:}
For data acquisition, judicious hardware and software design of the communication bus and protocol is essential to achieve optimal performance. Although the SPI bus can support clock frequencies up to $40~\mathrm{MHz}$, the actual speed may be impacted by various factors, such as electromagnetic interference (EMI)  and circuit routing. Adding more sensors can also lead to a reduction in transmission speed. Utilizing a self-designed flexible printed circuit with electromagnetic shielding, along with Low Voltage Differential Signalling (LVDS) routing, can significantly enhance performance. Additionally, implementing a COBS-based packeting protocol can improve the efficiency and reliability of data transmission. For data processing, we use multi-threading in Python to enable concurrent computing of tactile data from multiple sources within a quick response time. Different threads share the same memory space, alleviating the burden of computation-intensive executions and resource management, especially when handling multiple sources. Meanwhile, leveraging neural networks on the GPU helps to offload the CPU resources and mitigate issues such as race conditions during multi-processing. 

\textbf{Integration:}
While \acp{vbts} provide high spatial resolution and rich information, their bulkiness poses a challenge in relation to large-scale deployment. Our system tries to reduce the dimensions and weight of modular tactile sensors by leveraging customized image acquisition boards, modern micro-cameras, and lightweight external and internal supporting designs and materials. Additionally, by employing a cable-driven mechanism we also effectively reduce the overall weight and the dimensions of each finger for real-world grasping. This is notable in the case of multiple phalanges, as an alternative to installing motors at their joints.

\textbf{Calibration:}
In addition to addressing physical deployment challenges, we have streamlined the software complexity to simplify the calibration process, which can be laborious in real-world scenarios. By employing a modular sensor design, we have implemented a zero-shot calibration method that efficiently reduces calibration time and resources while maintaining high fidelity in tactile data collection.

\section{Conclusion \& Future Work}
\label{sec:conclusion}

This paper presented a systematic approach to overcome the difficulties inherent in the large-scale deployment of \acp{vbts}, which principally are data synchronization, ease of integration, and efficient calibration. A synchronized image acquisition system was first innovated for latency minimization. Based on that, a compact modular \ac{vbts} was then designed for ease of integration. Finally, a zero-shot calibration approach was proposed to avoid individual calibration, though without sacrificing accuracy.  For demonstration purposes, the overall system was deployed on a three-fingered robotic gripper with seven \acp{vbts} working simultaneously. By integrating the gripper into a robotic arm, the performance of the \acp{vbts} system was verified in both grasp and manipulation tasks.

\textbf{Future work} will focus on (i) improving the time efficiency and minimizing the power consumption of the synchronized image acquisition system; (ii) developing a realistic simulator for \acp{vbts}, to further facilitate their efficient calibration under varying conditions; (iii) generalizing this systematic approach to other vision-based tactile sensors for large-scale deployment; (iv) extending demonstrative deployment onto an anthropomorphic 5-fingered robotic hand with more than 15 \acp{vbts} integrated.

\textbf{Acknowledgement:} We thank Mr. Mish Toszeghi from QMUL for his meticulous proofreading of our manuscript. This work was supported in part by the National Natural Science Foundation of China (No. 62376031).

{
\small
\setstretch{0.95}
\bibliographystyle{ieeetr}
\bibliography{reference}
}
\end{document}